\documentclass[]{bytedance_seed}

\usepackage[toc,page,header]{appendix}

\usepackage[utf8]{inputenc} 
\usepackage[T1]{fontenc}    
\usepackage{hyperref}       
\usepackage{url}            
\usepackage{booktabs}       
\usepackage{amsfonts}       
\usepackage{nicefrac}       
\usepackage{microtype}      
\usepackage[dvipsnames,table]{xcolor}         
\usepackage{graphicx}
\usepackage{enumitem}
\usepackage{makecell}
\usepackage{amssymb}
\usepackage{caption}
\usepackage{subcaption}
\usepackage{bbding}
\usepackage{booktabs}
\usepackage{multirow}
\usepackage{tabularx}
\usepackage{array}
\usepackage{amsmath}
\usepackage{wrapfig}

\newlength\savewidth\newcommand\shline{\noalign{\global\savewidth\arrayrulewidth
  \global\arrayrulewidth 1pt}\hline\noalign{\global\arrayrulewidth\savewidth}}
\newcommand{\tablestyle}[2]{\setlength{\tabcolsep}{#1}\renewcommand{\arraystretch}{#2}\centering\footnotesize}
\renewcommand{\paragraph}[1]{\vspace{1.25mm}\noindent\textbf{#1}}

\newcolumntype{x}[1]{>{\centering\arraybackslash}p{#1pt}}
\newcolumntype{y}[1]{>{\raggedright\arraybackslash}p{#1pt}}
\newcolumntype{z}[1]{>{\raggedleft\arraybackslash}p{#1pt}}

\newcommand{\app}{\raise.17ex\hbox{$\scriptstyle\sim$}}

\definecolor{deemph}{gray}{0.6}

\definecolor{baselinecolor}{gray}{.9}

\usepackage[linesnumbered,ruled,vlined,algo2e]{algorithm2e}
\usepackage{amsmath}
\usepackage{xcolor}

\SetCommentSty{mycommfont}

\title{UniVR: Thinking in Visual Space for \\ Unified Visual Reasoning}

\author[1,2,*]{Zhongwei Ren}
\author[1]{Yunchao Wei}
\author[1]{Yao Zhao}
\author[2]{Weibo Gong}
\author[2]{Xiao Liu}
\author[2]{Anran Wang}
\author[2, *, \dagger]{Xiangtai Li}
\author[1, *]{Xiaojie Jin}

\affiliation[1]{Beijing Jiaotong University}
\affiliation[2]{ByteDance}

\contribution[*]{Equal Contribution}
\contribution[\dagger]{Project Lead}

\abstract{
Learning broad world knowledge directly from raw visual data is a fundamental capability of intelligence. We introduce \textbf{UniVR}, the first investigation into simultaneously learning complex reasoning, fine-grained physical dynamics, and long-term planning from pure visual demonstrations. At its core, UniVR features VR-GRPO, a reinforcement learning paradigm with complementary global and step-level rewards. This approach enforces logical coherence and physical consistency throughout the reasoning process without requiring task-specific heuristics or image-text pairs. To train and evaluate UniVR, we construct \textbf{VR-X}, a large-scale benchmark curated from 16 diverse sources spanning long-horizon manipulation, spatial puzzles, and physical reasoning. It is the first comprehensive suite to assess these heterogeneous capabilities under a purely visual protocol. Remarkably, UniVR achieves up to a 25\% improvement on VR-X, and its superior visual reasoning also boosts performance on various multimodal understanding benchmarks. These findings underscore the vast potential of reasoning within visual spaces, with all code, data, and models are open-sourced for further research.
}

\date{\today}
\correspondence{Xiaojiao Jin, Yunchao Wei}

\checkdata[Project Page]{\href{https://maverickren.github.io/UniVR.github.io/}{\texttt{https://UniVR.github.io/}}}

\begin{document}
\maketitle


\section{Introduction}

Current AI models primarily derive world knowledge from text~\cite{lin2025parm,gpt,shao2024deepseekmath,touvron2023llama,touvron2023llama2,bai2023qwen,bai2025qwen2}, performing reasoning~\cite{havrilla2024teaching,han2024roserevolutionizingopensetdense,kojima2022large,ren2024pixellm} and planning~\cite{huang2022planner,shi2024enhancing,song2023llm,wang2023describe} within the textual space. However, text is an abstract representation of the world, which is unable to fully encompass the rich information of the real visual world, such as complex dynamics, spatial relationships, and underlying physical laws. In contrast, vision serves as the most direct medium for world knowledge and remains the primary source for animals and humans to acquire information. In most scenarios, humans can perform complex reasoning without relying on language by directly simulating task execution and scene transitions in their minds, which constitutes our innate visual reasoning capability. Given the vast abundance of video content available on the internet, equipping AI with the capacity for complex reasoning and planning within the visual space holds significant promise for enhancing its world modeling abilities and pushing the frontier of efficient task execution in the real visual world.

\begin{figure}
    \centering
    \includegraphics[width=\linewidth]{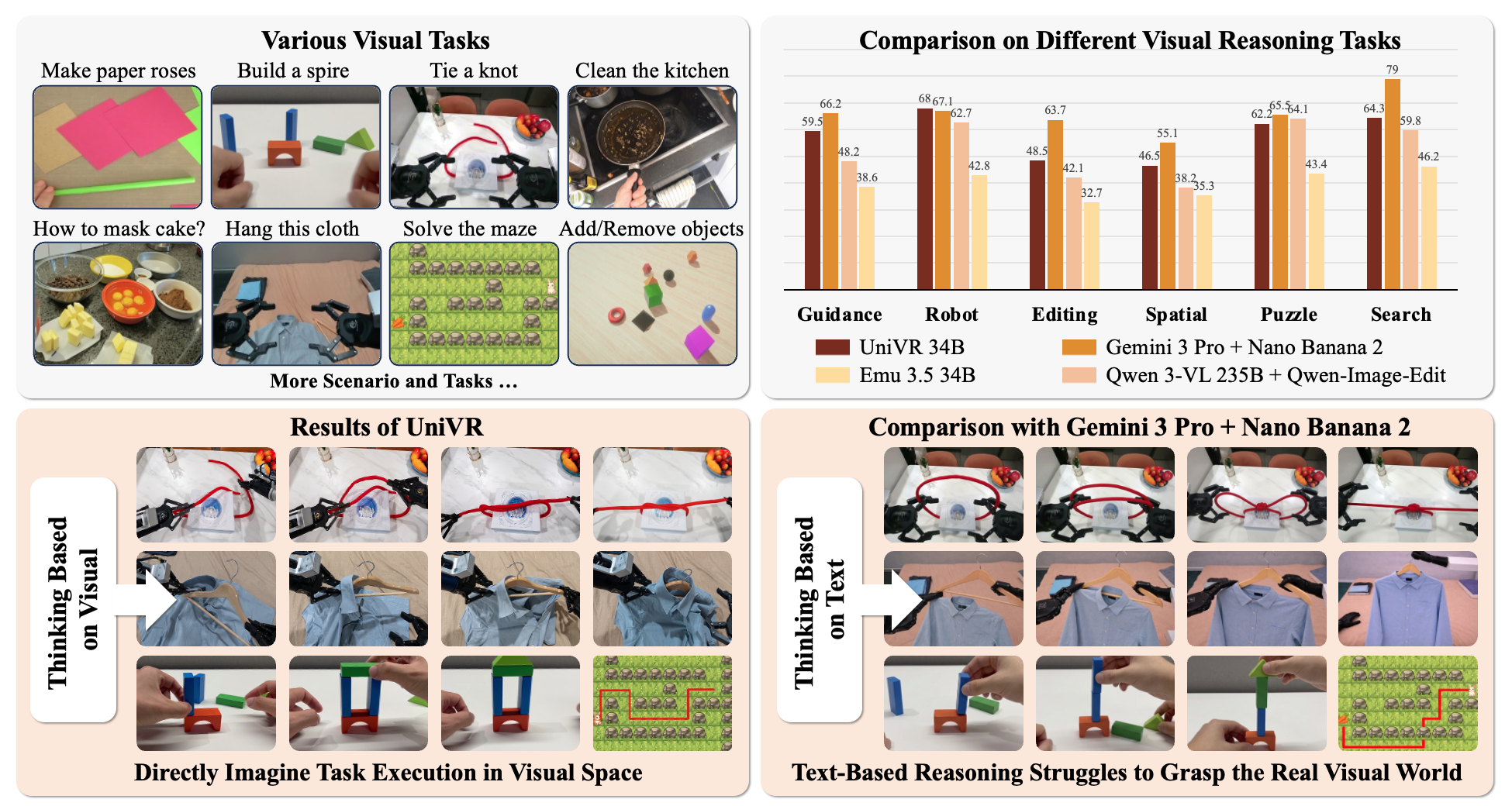}
    \caption{UniVR focuses on advancing native visual-space reasoning and planning across diverse tasks. Compared to LMMs that reason within textual space, UniVR facilitates a more profound comprehension of real  visual world, improving policy learning in various scenarios.}
    \label{fig:fig1}
\end{figure}

Recent research has explored two primary pathways for advancing visual reasoning. On one hand, MLLMs, such as Gemini-3~\cite{gemini3} and GPT-5~\cite{singh2025openai}, can generate detailed textual reasoning chains that are subsequently converted into visual reasoning traces via generative models like Nano Banana~\cite{comanici2025gemini} and GPT-Image~\cite{openai2025gptimage1}. While this paradigm conveniently leverages pre-existing linguistic strengths, textual abstractions inherently struggle to capture the intricate dynamics and spatial relationships of the physical world. Such limitations prevent a seamless transfer of textual reasoning proficiency into the visual domain. As shown in Fig~\ref{fig:fig1}, even with state-of-the-art reasoning and high-fidelity rendering, models frequently fail to maintain logical coherence and physical consistency in tasks requiring fine-grained, long-horizon visual evolution. 
On the other hand, some works~\cite{ren2026videoworld, ren2025videoworld,tan2026dreamworld,mialon2023gaia,russell2025gaia} leverage video generation models to model task strategies directly in visual space, while others employ unified generation models~\cite{cui2025emu3,wang2024emu3,xie2025show,deng2025emerging,wu2025omnigen2} for more comprehensive reasoning through joint vision-language representations. Despite their promise in acquiring complex knowledge from visual data, their reasoning still remains heavily dependent on text guidance, requiring dense image-text pairs for both training and inference. These limitations constrain the scaling of visual reasoning and drive our core inquiry: \textbf{How can models utilize raw visual data to boost their visual reasoning capabilities across diverse tasks?}

To investigate this problem, we first establish VR-X, a comprehensive benchmark designed to assess visual reasoning capabilities across diverse tasks. As shown in Fig~\ref{fig:vrx}, VR-X encompasses two primary task categories. The first focuses on long-horizon complex planning across various environments, featuring minute-scale tasks with fine-grained dynamics (\textit{e.g.} tying knots, folding clothes) and diverse domains spanning cooking, crafts, robotic control, and navigation.
The second focuses on general visual reasoning, evaluating fundamental cognitive skills such as visual search, puzzles, spatial perception, and editing. Models are tasked to directly reason and generate in visual space, with assessment centered on the logical coherence and task completion of the visual reasoning trace. Such task diversity and evaluation paradigm is rarely explored in existing benchmarks, demanding robust learning capabilities to master heterogeneous task knowledge.

Based on this benchmark, we evaluate state-of-the-art MLLMs, including Gemini~\cite{gemini3,comanici2025gemini}, GPT~\cite{singh2025openai}, and Qwen~\cite{qwen3.5,bai2025qwen3vl}, by prompting them to generate detailed textual reasoning chains to guide generative models, alongside visual generation models such as Emu3.5~\cite{cui2025emu3}. Results indicate that while they excel at textual reasoning, visual comprehension, or high-fidelity generation, they still struggle to accurately execute tasks in the benchmark. As shown in Fig.~\ref{fig:fig1} and Fig.~\ref{fig:failure_case}, their predictions contain errors such as logical gaps, physical inconsistency, and violations of rules, ultimately failing to produce visually coherent task sequences. This underscores an urgent need for frameworks that effectively master those heterogeneous knowledge.

Motivated by these observations, we propose \textbf{UniVR}, a unified next-token prediction framework with strong visual reasoning capabilities. At its core is VR-GRPO, a novel RL paradigm that learns diverse knowledge directly in visual space. A VLM evaluator first provides holistic assessment of generated sequences for task completion and visual quality. However, as detailed in Sec.~\ref{subsec:vrgrpo}, this vanilla reward fails to capture errors like logical gaps and physical inconsistency.  
 We attribute this to current VLMs' limited visual world knowledge to identify fine-grained errors in minute-level multi-step sequences, leaving final states and appearance quality to dominate judgments. We therefore propose a \textbf{Step-Focal} reward that can proactively targets error-prone substeps for more precise assessment. Combined with global evaluation, this design ensures both overall task completion and fine-grained reasoning coherence, without relying on dense image-text pairs or task-specific rules.

In addition to these qualitative analyses, we benchmark the performance of UniVR against existing methods on \textbf{VR-X}. As shown in Fig.~\ref{fig:fig1}, UniVR significantly boosts visual reasoning capabilities without compromising the foundational strengths of the base model (Emu3.5). Notably, with only 34B parameters, UniVR approaches the performance of the Gemini 3 Pro~\cite{gemini3} + Nano Banana 2~\cite{google2025gemini25flashimage} pipeline and even surpasses Gemini 3 in long-horizon manipulation tasks. This demonstrates the superior efficiency and effectiveness of our visual reasoning framework. Further visualization results are provided in Fig.~\ref{fig:more_vis} to demonstrate its robust performance across diverse scenarios.

Our contributions are summarized as follows:
\begin{itemize}[leftmargin=*]
    \item \noindent We are the first to explore learning heterogeneous tasks, from long-term planning to general cognitive reasoning, directly in a unified visual space without language supervision.
    \item \noindent We propose VR-GRPO, featuring a novel Step-Focal reward that proactively targets error-prone reasoning substeps alongside a global reward. This significantly improves logical coherence and physical consistency in visual reasoning without relying on image-text pairs or task-specific rules.
    \item \noindent We introduce VR-X, a benchmark for evaluating diverse tasks in visual space, spanning from fine-grained long-term planning to general reasoning, to facilitate future research on visual reasoning.
    
\end{itemize}
\section{Related Works}

\noindent \textbf{Visual Reasoning.}
The rapid advancement of LLMs~\cite{chowdhery2023palm, gpt,bai2023qwen,yang2024qwen2,jiang2023mistral,wang2024world,yang2025qwen3} and MLLMs~\cite{bai2025qwen2,wu2024ivideogpt, liu2023visual,bai2025qwen3vl,team2026qwen3omni} has spurred extensive research on text-centric reasoning paradigms, such as chain-of-thought~\cite{zhang2022automatic, wei2022cot,madaan2022text} and latent-space reasoning~\cite{hao2024coconut,chen2025reasoning}. 
Recent efforts like visual CoT~\cite{li2025latent,wang2025monet,qin2025chain} extend this paradigm to multimodal inputs, yet they still project visual features into a linguistic space, leaving reasoning fundamentally text-bounded . 
Another line of work explores video generation for non-linguistic world modeling in autonomous driving~\cite{mialon2023gaia,russell2025gaia} and robotics~\cite{tan2026dreamworld,guo2025ctrl}. 
However, these approaches are typically confined to short-horizon dynamics or narrow, single-task settings. 
UniVR departs from both paradigms by directly reasoning in the visual space, unifying and  improving heterogeneous tasks, which demand long-horizon planning, complex policy and intricate physics.

\noindent \textbf{Unified Model}~\cite{li2025onecat,deng2025emerging,huang2025illume+,xie2025show,wu2025omnigen2,qin2025star,yang2025mmada} aims to encode world knowledge in a single model space, combining the textual reasoning of MLLMs with the visual generation capabilities.
Pioneering architectures~\cite{cui2025emu3,wang2024emu3} tokenize images, text, and video into a unified discrete space to enable flexible cross-modal generation. 
However, these models are predominantly trained on entertainment-oriented or artistic editing objectives, and their visual knowledge acquisition remains heavily constrained by dense image-text supervision. 
Our UniVR breaks this dependency by incorporating a text-free visual reasoning framework into the unified architecture, enabling the model to learn complex reasoning and planning directly from raw visual inputs.

\noindent \textbf{Reinforcement Learning in Generative Model.} The success of RL in MLLMs~\cite{schulman2017ppo,rafailov2023dpo,ouyang2022rlhf,feng2025videor1,yang2025r1onevision,lin2025parm,yu2025dapo,tan2025reasonrft,liu2025segzero} has been extended to visual generation, with pioneering works such as DDPO~\cite{black2023ddpo} and ReFL~\cite{xu2023imagereward} employing PPO~\cite{schulman2017ppo} or RLHF~\cite{ouyang2022rlhf} to align models with human preferences regarding image fidelity. 
Following the success of DeepSeek-R1~\cite{guo2025deepseek}, GRPO~\cite{shao2024deepseekmath,liu2025flowgrpo} has been widely adopted across various vision tasks to enhance multimodal understanding and image/video generation. 
Various approaches~\cite{wang2025unified,xue2025dancegrpo,yuan2025ar} also incorporate specialized reward, such as boosting aesthetic appeal via HPSv3~\cite{ma2025hpsv3,wu2023hpsv2} or enforcing semantic alignment through CLIP~\cite{radford2021clip} and VLM-based scoring. 
However, these reward mechanisms primarily target perceptual quality, text-image consistency, or single-step correctness. 
In contrast, our VR-GRPO is specifically designed to optimize for logical consistency and physical plausibility in the context of multi-step visual reasoning and policy learning.
\section{UniVR}

\begin{figure}
    \centering
    \includegraphics[width=\linewidth]{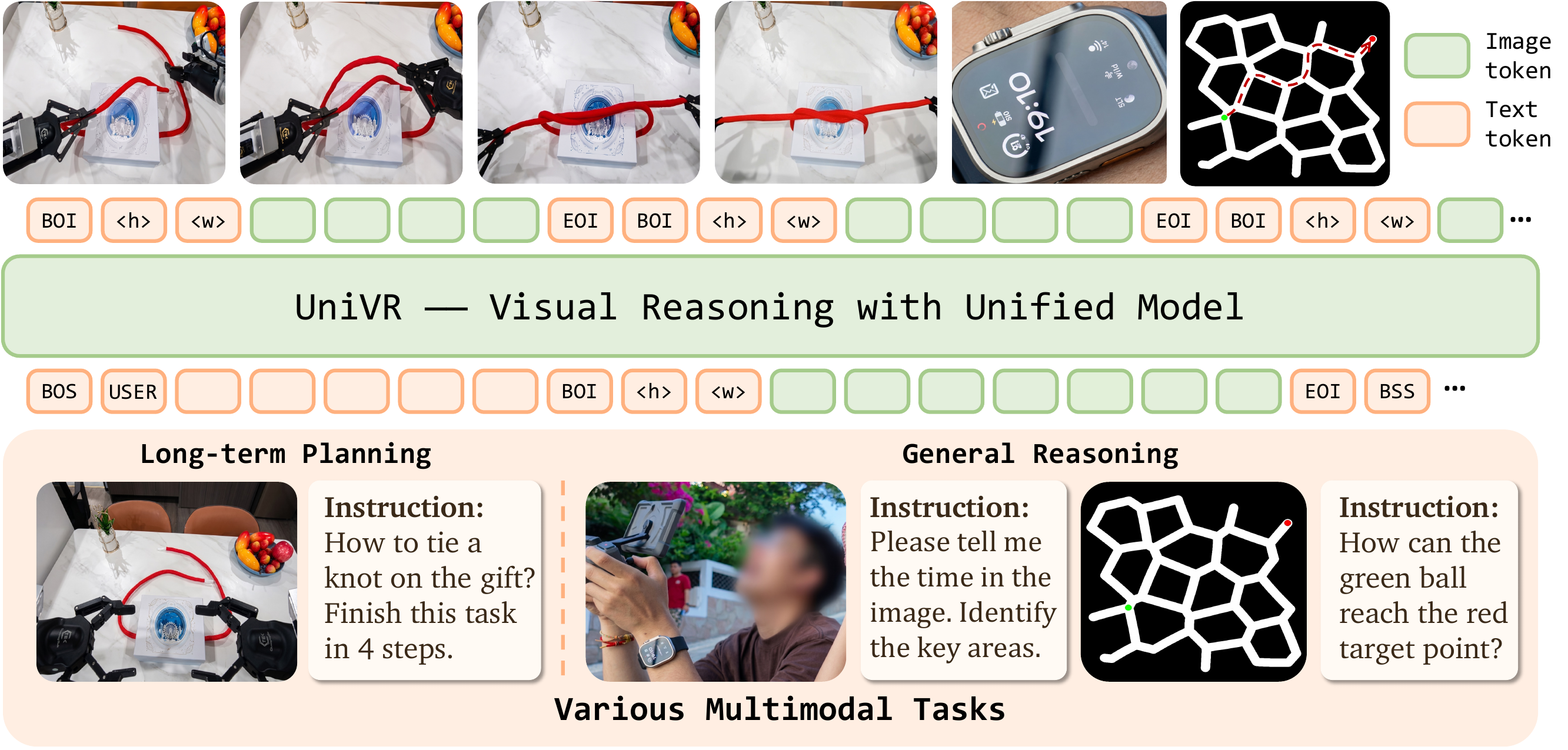}
    \caption{\textbf{Overview of UniVR architecture.} Via a unified next-token prediction objective, UniVR processes instructions and image queries to directly generate visual reasoning traces for task execution.}
    \label{fig:method}
\end{figure}

In this section, we introduce UniVR, shown in Fig.~\ref{fig:method}, which adopts an autoregressive generation model as its basic framework. It has a two-stage training pipeline: cold initialization and reinforcement learning. We describe how to conduct visual space reasoning using this framework, together with cold initialization in Sec.~\ref{subsec:cold}, and RL training in Sec.~\ref{subsec:vrgrpo}.

\subsection{Thinking in Visual Space}
\label{subsec:cold}
Given an image sequence $x_{1:t}$ and an instruction, we model the next-frame distribution: $p(x_{t+1} \mid x_{1:t})$. We use image sequences containing demonstration trajectories of task execution across diverse scenarios, encompassing various kinds of planning and reasoning knowledge. Our formulation does not require dense textual reasoning chains, but instead directly models the state transitions and underlying policy dynamics in these trajectories, encouraging the model to reason in visual space.

Specifically, we adopt Emu3.5 as our baseline, a state-of-the-art unified generative model that produces variable-length image sequences. It employs a VQ-VAE-style autoencoder~\cite{oord2018vqvae} to encode images and text into a unified discrete vocabulary, also enabling text generation. This allows us to further investigate how enhanced visual reasoning affects multimodal understanding (See Sec.~\ref{subsec:abl}). 
Despite pretraining on extensive image-text pairs, this baseline still struggles to capture complex real-world physics, fine-grained action dynamics, and visual cognitive abilities, as shown in Sec.~\ref{subsec:mainres} and Fig.~\ref{fig:failure_case}. We attribute this to its heavy reliance on dense textual reasoning chains for world knowledge acquisition, reducing vision to a mere renderer. 
We therefore first perform supervised fine-tuning on a curated dataset of diverse visual tasks as cold initialization, standardizing all samples as {query image, instruction, visual reasoning trajectory}. This endows the model with visual reasoning priors for subsequent RL. Training configuration and data construction are detailed in Appendix~\ref{sec:appA}.

\subsection{Visual Reasoning GRPO}
\label{subsec:vrgrpo}

Despite the versatility of casting various tasks into a unified visual space, we observe that vanilla SFT still struggles to reconcile heterogeneous multi-source tasks. For instance, 2D puzzles and long-term human planning ediffer drastically in temporal scale, domain knowledge, and visual appearance.
As detailed in Sec.~\ref{subsec:abl}, dense supervision in SFT training fails to yield consistent improvements across diverse tasks. 
While RL method, \textit{e.g.} GRPO~\cite{shao2024deepseekmath}, offer a promising paradigm for performance enhancement by enabling models to autonomously explore optimal policies across diverse scenarios, existing methods prioritize visual fidelity or cross-modal alignment rather than visual reasoning. 
\begin{figure}
    \centering
    \includegraphics[width=\linewidth]{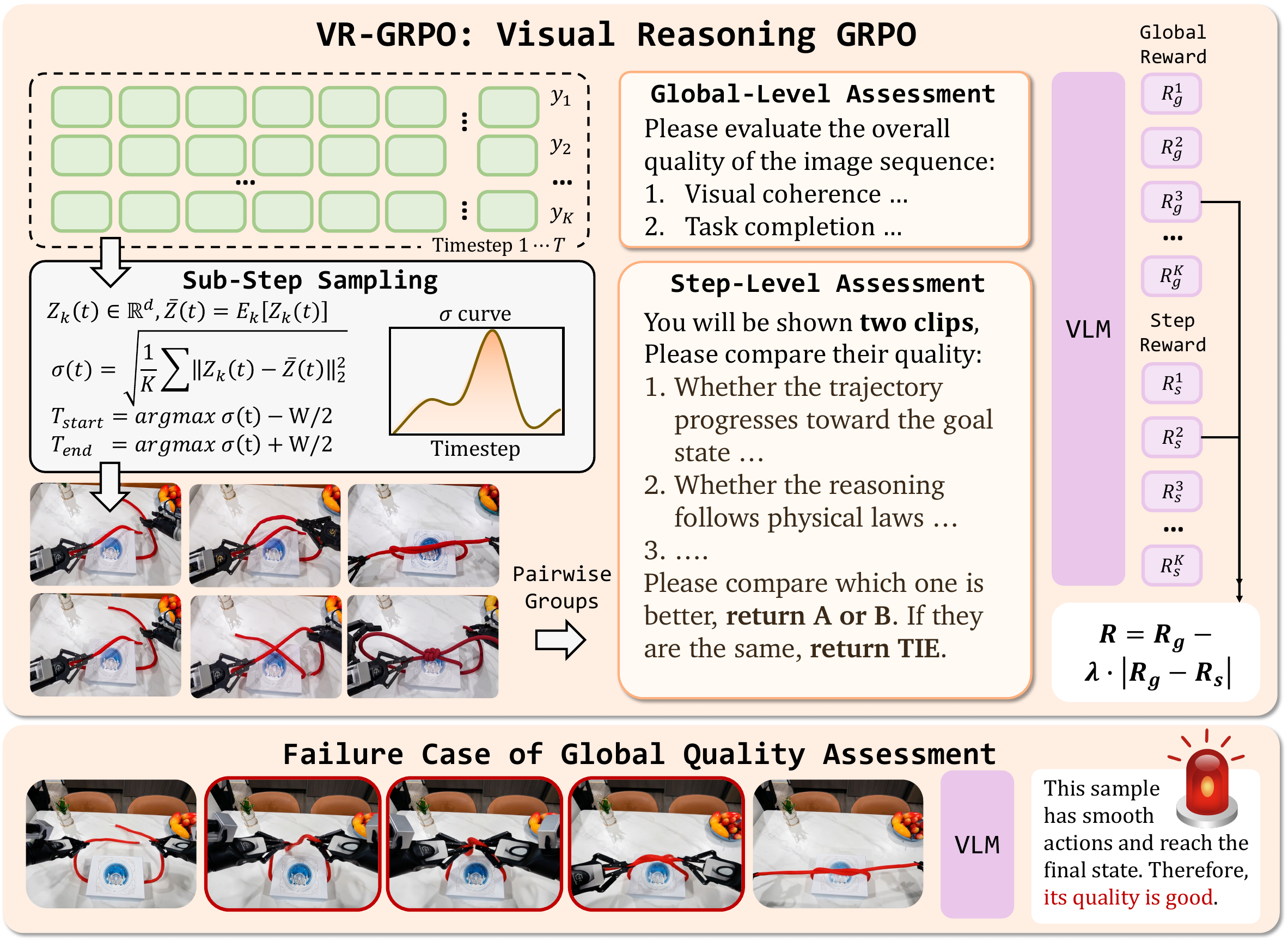}
    \caption{\textbf{The proposed Visual Reasoning GRPO. } (up) VR-GRPO integrates global and step-focal rewards to ensure both task completion and the physical coherence of generated reasoning traces. (down) Failure case of global rewards that overlook local inconsistencies within long reasoning traces.}
    \label{fig:vrgrpo}
\end{figure}

To address this, we introduce VR-GRPO, an RL methodology that eschews the requirement for dense image-text pairs or task-specific heuristics, focusing on logical coherence and task completion. 
In Sec.~\ref{subsec:abl}, we further demonstrate that VR-GRPO can also seamlessly integrate with textual reasoning tasks, facilitating a synergistic advancement in overall multimodal understanding.

\noindent \textbf{Reward design.}
The VR-GRPO has two reward components: format reward $R_{\rm format}$ and visual reasoning reward $R_{\rm reason}$. $R_{\rm reason}$ consists of a global reward $R_{\rm g}$
and a novel step-focal reward $R_{\rm s}$.
$R_{\rm format}$ ensures that the generated image sequences satisfy structural constraints, such as uniform resolution and the number of reasoning steps prescribed by the task instructions.  

For $R_{\rm reason}$, we use a general-purpose prompt to guide an online reward model (\textit{i.e.} Qwen3-VL-30B) in providing an overall quality assessment of rollout samples against ground-truth references.  
While the VLM effectively assesses task completion and visual fidelity, as shown in Fig.~\ref{fig:reward_comp}, our preliminary analysis indicates that this global reward often overlooks intermediate physical violations and logical gaps, over-prioritizing terminal success and pixel clarity, especially in long-horizon, multi-step tasks. We attribute this to current VLMs' reliance on text-derived world knowledge, which constrains their ability to pinpoint fine-grained visual dynamic errors in minute-level multi-step sequences.



Consequently, we introduce a step-focal reasoning reward $R_s$ designed to identify the most error-prone steps in the reasoning trajectory. 
By focusing the VLM on these critical steps, we deliver precise, fine-grained rewards to ensure the model maintains logical and physical coherence throughout complex, long-horizon tasks. 
Specifically, for a set of $K$ rollout trajectories ${y_{1:K}}$, we first assume a uniform length $T$. 
We employ a CLIP image encoder to extract per-frame feature embeddings as $z_k(t)\in\mathbb{R}^d$ and calculate the inter-trajectory variance at each timestep $t$ as:
\begin{equation}
    \sigma(t) = \sqrt{\frac{1}{K} \sum_{k=1}^K \|\mathbf{z}_k(t) - \bar{\mathbf{z}}(t)\|_2^2}
\end{equation}
where $\bar{\mathbf{z}}(t)$ is the mean embedding across all trajectories at time $t$. 
A high variance $\sigma(t)$ indicates a state of maximum uncertainty where the model's reasoning paths diverge. 
Given a window size $W$, we identify the peak uncertainty at $t^* = \text{argmax}_t \sigma(t)$ and extract a reasoning segment from $[t^* - W/2, t^* + W/2]$. 
For samples that are excessively challenging, this process reverts to random sampling. In practice, generated trajectories often vary in length. 
To ensure temporal alignment, we partition each trajectory into an equal number of segments. We then treat the segment index as the synchronized timestep and use the average CLIP features within each segment to perform the uncertainty analysis described above. 
Following~\cite{wang2025pref}, we employ a pairwise evaluation protocol for both global rewards and the identified sub-steps to derive relative win rates. 
This approach is designed to further mitigate the inherent scoring bias that often accompanies direct VLM-based scalar assessments. 
Upon obtaining the global score $R_{\rm g}$ and step-focal scores $R_{\rm s}$ for each trajectory, we integrate them using the following formula:
\begin{equation}
    R_{\rm reason} = R_{\rm g} - \lambda |R_{\rm g} - R_{\rm s}|
\end{equation}
This formulation prevents the model from taking reasoning shortcuts, as it necessitates that the model not only predicts the terminal state accurately but also maintains procedural integrity and physical coherence. 
The coefficient $\lambda$ controls the strength of this alignment.

\section{VR-X Benchmark}

Our objective is to advance the capacity for reasoning natively within the visual domain across a wide array of scenarios. We prioritize a comprehensive benchmark capable of evaluating multi-step planning, delicate manipulations and cognitive reasoning. However, current benchmarks mostly emphasize visual quality or text-matching, limited to restricted environments and short-term tasks. 

\begin{figure}
    \centering
    \includegraphics[width=\linewidth]{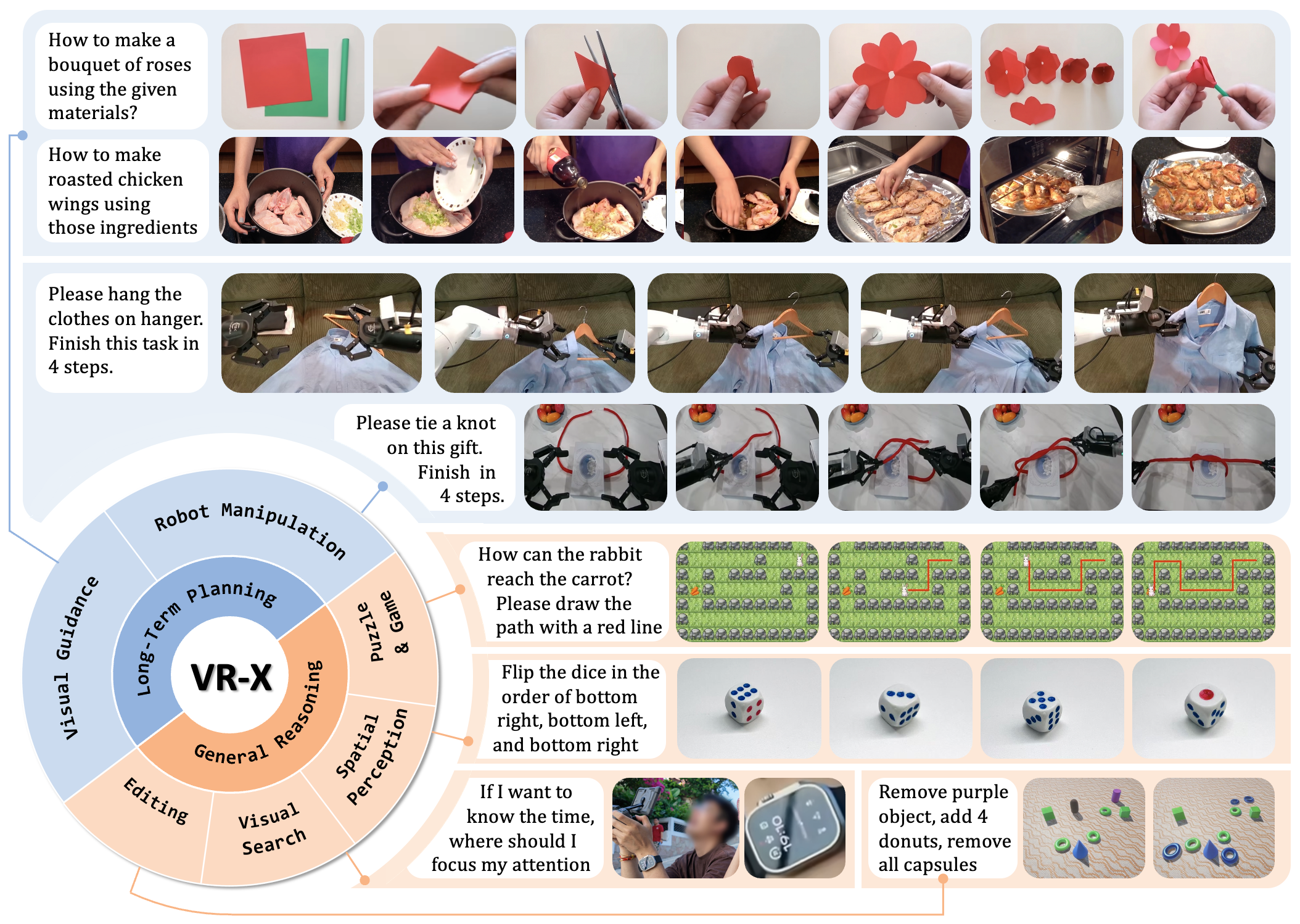}
    \caption{\textbf{Overview of VR-X benchmark.}  }
    \label{fig:vrx}
\end{figure}

Therefore, we introduce VR-X, the first large-scale benchmark designed for diverse and heterogeneous visual reasoning. As shown in Fig.~\ref{fig:vrx}, it includes six tasks, each providing detailed visual reasoning traces: visual guidance, robotic manipulation, puzzle games, editing, search, and spatial perception.

\subsection{Dataset Generation}
VR-X is characterized by its vast diversity and involves fine-grained, visually complex manipulations that are hard to describe linguistically. We curate 1.5M raw samples from 16 diverse sources (e.g., AgiBot~\cite{bu2025agibot}, Action100M~\cite{chen2026action100m}, EgoDex~\cite{hoque2025egodex}, VisualCoT~\cite{shao2024visual}), spanning minute-long planning (robotic manipulation, cooking, handcrafting) to single-step reasoning (mazes, visual search). Rigorously curated into 310k cold-start training, 3k RL, and 1.8k benchmark evaluation samples, all follow a unified format: {query image, textual instruction, visual reasoning trajectory}. We further annotate these sequences with fine-grained textual chain-of-thought (CoT) descriptions to support multimodal learning. See Appendix~\ref{sec:appA} for more details.



\subsection{Evaluation Metrics}

Our evaluation focuses on two main aspects: the logical accuracy of visual reasoning and the adherence to real physical dynamics. 
Accordingly, we employ the following metrics for evaluation:

\noindent \textbf{VLM score:} We conduct an automated evaluation using Qwen3.5-397B~\cite{qwen3.5} to quantitatively assess reasoning traces. The model evaluates each sample based on task completion, procedural coherence, visual informativeness, and image fidelity. These dimensions are aggregated into a final normalized score (0–100), providing a robust metric for measuring the logical and visual quality of the results.

\noindent \textbf{JEPA similarity:} Despite strong reasoning capabilities, VLMs' dependency on textual knowledge often hinders their ability to perceive intrinsic physical laws. As shown in Fig.~\ref{fig:reward_comp}, this may result in hallucinations where the evaluator ignores fine-grained physical inconsistencies. To bridge this gap, we incorporate JEPA similarity as an additional metric. Building on prior research~\cite{luo2024beyond} demonstrating that V-JEPA~\cite{assran2025vjepa2} encoders can capture high-level latent physical dynamics, we map sequences into latent space and compute maximum mean discrepancy against ground-truth distributions. Lower scores indicate closer alignment with real-world physics. See Appendix~\ref{sec:appA} for more evaluation details.

\section{Experiment}
\label{sec:exp}
\begin{table}[t]
\centering
\caption{\textbf{Comparison on VR-X.} Qwen, Gemini 2.5/3, and GPT-5 are paired with Qwen-image-edit, Nano Banana 1/2, and GPT-image-1.5, respectively.  }
\vspace{0.5em}
\resizebox{\textwidth}{!}{%
\begin{tabular}{lccccccccc}
\toprule
\multirow{2}{*}{Method} & \multirow{2}{*}{\makecell{Visual\\Thinking}} & \multicolumn{2}{c}{Long-term planning} & \multicolumn{4}{c}{General Reasoning} & \multirow{2}{*}{Overall$\uparrow$} & \multirow{2}{*}{JEPA$\downarrow$} \\
\cmidrule(lr){3-4} \cmidrule(lr){5-8}
                        & & Guidance & Robot & Editing & Spatial & Puzzle & Search & & \\
\midrule
\multicolumn{10}{l}{{\color{gray}\textit{Large Multimodal Model + T2I Model}}} \\
 Qwen3-VL-235B~\cite{bai2025qwen3vl}  &\small{\XSolidBrush} &48.2 &62.8 &42.1 &38.2 &64.1 &59.8 &52.5 &18.08 \\
 Qwen3.5-397B~\cite{qwen3.5}  &\small{\XSolidBrush} &47.0 &63.2 &39.8 &40.4 &\underline{65.6} &64.5 &53.4 &18.64  \\
GPT-5~\cite{singh2025openai} &\small{\XSolidBrush} &\textbf{68.2} &64.1 &\underline{58.0} &\underline{49.3} &64.0 &\underline{77.4} &\underline{63.5} &\underline{12.17} \\
 Gemini-2.5-pro~\cite{comanici2025gemini} &\small{\XSolidBrush} &58.4 &\textbf{67.9} &54.0 &40.5 &\textbf{67.7} &76.3 &60.8 &14.39 \\
Gemini-3-pro~\cite{gemini3} &\small{\XSolidBrush} &\underline{66.2} &\underline{67.1} &\textbf{63.7} &\textbf{55.1} &65.5 &\textbf{79.0} &\textbf{66.1} &\textbf{11.07} \\

 \midrule


 \multicolumn{10}{l}{{\color{gray}\textit{Unified Generation Model}}} \\
 Janus-pro~\cite{chen2025janus} &\small{\XSolidBrush} &9.2 &18.2 &5.4 &10.2 &27.1 &21.5 &15.3 &68.79 \\
 Show-o2~\cite{xie2025show} &\small{\XSolidBrush} &15.1 &22.5 &13.0 &17.1 &29.4 &35.8 &22.2 &59.93 \\
 ILLUME+~\cite{huang2025illume+} &\small{\XSolidBrush} &13.1 &11.5 &5.8 &14.6 &22.2 &27.5 &15.8 &61.12 \\
 OneCAT~\cite{li2025onecat} &\small{\XSolidBrush} &15.6 &13.5 &10.3 &20.2 &16.2 &30.1 &17.7 &77.06 \\
 STAR~\cite{qin2025star} &\small{\XSolidBrush} &22.7 &28.5 &14.2 &\underline{21.5} &27.2 &37.7 &25.3 &51.98 \\
 OmniGen2~\cite{wu2025omnigen2} &\small{\XSolidBrush} &20.4 &29.4 &15.2 &16.9 &30.5 &42.1 &25.8 &47.09 \\
 Bagel~\cite{deng2025emerging} &\small{\XSolidBrush} &25.2 &34.7 &20.9 & 21.3 &35.1 &\underline{47.7} &30.8 &40.88 \\
 Emu3.5~\cite{cui2025emu3} &\small{\XSolidBrush} &\underline{38.6} &\underline{42.8} &\underline{32.7} &35.3 &\underline{43.4} &46.2 &\underline{39.8} &\underline{33.62} \\

 \midrule
 UniVR &\small{\Checkmark} &\textbf{59.5} &\textbf{68.0} &\textbf{48.5} &\textbf{46.5} &\textbf{62.2} &\textbf{64.3} &\textbf{58.2} &\textbf{13.01} \\

 $\triangle$ \textit{v.s.} Emu 3.5 & &\textcolor{ForestGreen}{$\uparrow$20.9} &\textcolor{ForestGreen}{$\uparrow$25.2} &\textcolor{ForestGreen}{$\uparrow$15.8} &\textcolor{ForestGreen}{$\uparrow$11.2} &\textcolor{ForestGreen}{$\uparrow$18.8} &\textcolor{ForestGreen}{$\uparrow$18.1} &\textcolor{ForestGreen}{$\uparrow$18.4} &\textcolor{ForestGreen}{$\downarrow$20.61} \\

\bottomrule
\end{tabular}%
}
\label{tab:tab1}
\end{table}

We leverage Emu3.5 34B to initialize our unified model, which undergoes full-parameter SFT and RL. Our RL pipeline is implemented using the verl framework~\cite{sheng2025hybridflow}, utilizing a rollout size of 8. For sub-step selection, we set the default window size to 4 frames and the consistency coefficient $\lambda$ to 2.0. During training, video resolution is scaled to a short-side of 512px, and the maximum sequence length is capped at 20k tokens. See Appendix~\ref{sec:appA} for more implementation details.

\subsection{Results on VR-X}
\label{subsec:mainres}
Tab.~\ref{tab:tab1} provides a comprehensive comparison on the VR-X benchmark across two dominant technical paradigms: the integration of large multimodal models with T2I models and unified generation models. This evaluation aims to investigate the capacity of current approaches for visual-space reasoning while validating the efficacy of UniVR.

\noindent \textbf{LMMs with T2I models.}
We first evaluate off-the-shelf LMMs (rows 1-5), including Qwen 3-VL~\cite{bai2025qwen3vl} and Qwen 3.5~\cite{qwen3.5} with Qwen-Image-Edit~\cite{wu2025qwenimagetechnicalreport}, Gemini 2.5/3 Pro~\cite{gemini3,comanici2025gemini} with Nano Banana 1/2~\cite{google2025gemini25flashimage,google2026nanobanana2}, and GPT-5~\cite{singh2025openai} with GPT-image 1.5~\cite{openai2025gptimage1}. These systems follow a two-stage pipeline: the LMM first produces step-level textual instructions from the input image and task prompt. The generation module then renders the sequence frame-by-frame conditioned on previous frames and textual guidance.
Gemini 3 Pro with Nano Banana 2 achieves the best performance, benefiting from strong textual reasoning and superior rendering quality. However, as shown in Fig.~\ref{fig:failure_case}, visual inconsistencies persist in tasks with complex dynamics and planning, even with detailed instructions. 
Notably, this issue does not improve substantially with the evolution of language models, comparisons between Gemini 3/2.5 Pro and Qwen 3.5/3-VL show marginal gains in visual logicality. This underscores the urgent need for intrinsic visual world knowledge beyond mere linguistic scaling.

\begin{table*}[t]
\centering
\subfloat[
Results on understanding benchmarks. $^*$ means training with with interleaved image-text sequences.
\label{tab:abla_understand}
]{
\begin{minipage}[t]{\linewidth}
\centering
\tablestyle{4pt}{1.08}
\footnotesize
\begin{tabular}{l| c c c c c c}
    Method & MMMU~\cite{yue2024mmmu}  & MME(P)~\cite{fu2023mme} &MME(C)~\cite{fu2023mme} & MMBench~\cite{liu2024mmbench} & MathVista~\cite{lu2023mathvista} & MM-Vet~\cite{yu2023mm} \\
    \shline
    \specialrule{0em}{0pt}{1pt}
     Emu 3.5    &\underline{0.292}  &781.1 &\underline{324.6} &0.183 &\underline{41.7} &28.0 \\
     Text-only training &0.290  &\underline{782.0} &323.4 &\textbf{0.199} &40.8 &\underline{28.3} \\
     UniVR$^*$  &\textbf{0.337}  &\textbf{799.3} &\textbf{338.5} &\underline{0.198} &\textbf{44.0} &\textbf{35.6}     \\
     $\triangle \textit{v.s.}$ Emu3.5 & \textcolor{ForestGreen}{$\uparrow$ 0.045} & \textcolor{ForestGreen}{$\uparrow$ 18.2} & \textcolor{ForestGreen}{$\uparrow$ 13.9} & \textcolor{ForestGreen}{$\uparrow$ 0.015} & \textcolor{ForestGreen}{$\uparrow$ 2.3} & \textcolor{ForestGreen}{$\uparrow$ 7.6} \\

\end{tabular}
\end{minipage}
}
\\
\vspace{0.55em}
\noindent\makebox[\linewidth][c]{%
\subfloat[
Reward in VR-GRPO.
\label{tab:abla_vrgrpo}
]{%
\begin{minipage}[t]{0.41\linewidth}
\centering
\tablestyle{3pt}{1.08} 
\footnotesize
\begin{tabular}{c c c c c c c}
    \multicolumn{1}{c}{\multirow{2}*{\makecell{Cold\\Start}}} & \multicolumn{3}{c|}{Reward} &\multicolumn{3}{c}{VR-X}    \\
      & Global &Step &\multicolumn{1}{c|}{Pairwise}  &LP &GR & JEPA$\downarrow$   \\
    \shline
    \specialrule{0em}{0pt}{1pt}
    \small{\checkmark} & & &\multicolumn{1}{c|}{} &48.2 &42.4 &18.44 \\
    \small{\checkmark} & \small{\checkmark} & &\multicolumn{1}{c|}{}  &45.7 &46.0 &22.30 \\
    \small{\checkmark} & &\small{\checkmark} & \multicolumn{1}{c|}{}  &54.9 &45.8 &15.87 \\
    \small{\checkmark} & \small{\checkmark} & \small{\checkmark} & \multicolumn{1}{c|}{} &\underline{61.6} &\underline{53.7} &\textbf{12.89} \\
    \small{\checkmark} & \small{\checkmark} & \small{\checkmark} & \multicolumn{1}{c|}{\small{\checkmark}}  &\textbf{63.8} &\textbf{55.4} &\underline{13.01} \\
\end{tabular}
\end{minipage}
}%
\hspace{0.03\linewidth}%
\subfloat[
Training with text-based RL method.
\label{tab:abla_modality}
]{%
\begin{minipage}[t]{0.52\linewidth}
\centering
\tablestyle{3pt}{1.08}
\footnotesize
\begin{tabular}{c c c c c| c c c}
      \multicolumn{1}{c}{\multirow{2}*{\makecell{Training \\ Data }}} &\multicolumn{1}{c}{\multirow{2}*{\makecell{Cold\\Start}}} &\multicolumn{3}{c|}{Reward} &\multicolumn{3}{c}{VR-X}\\
       & \multicolumn{1}{c}{} & HPSv3 & CLIP & VR-GRPO &LP &GR & JEPA$\downarrow$  \\
    \shline
    \specialrule{0em}{0pt}{1pt}
      \multicolumn{1}{c}{V} & \small{\checkmark} & & &  &48.2 &\underline{42.4} &18.44 \\
    \multicolumn{1}{c}{V\&T} & \small{\checkmark} & &  & &49.7 &41.0  &18.37 \\
   \multicolumn{1}{c}{V\&T}  & \small{\checkmark} &\small{\checkmark} & & &49.0 &41.7 &18.09 \\
     \multicolumn{1}{c}{V\&T} & \small{\checkmark} &\small{\checkmark} &\small{\checkmark} & &\underline{52.6} &42.0 &\underline{17.17}  \\
     \multicolumn{1}{c}{V\&T} & \small{\checkmark} &\small{\checkmark} &\small{\checkmark} &\small{\checkmark} &\textbf{65.4} &\textbf{57.8} &\textbf{12.97}\\
\end{tabular}
\end{minipage}
}%
}
\\
\vspace{0.55em}
\hfill
\subfloat[
Ablation on joint training.
\label{tab:abla_heter}
]{
\begin{minipage}[t]{0.27\linewidth}
\centering
\tablestyle{4pt}{1.08}
\footnotesize
\begin{tabular}{l| c c c}
    Method & Training Data & LP & GR \\
    \shline
    \specialrule{0em}{0pt}{1pt}
      \rowcolor{gray!20}  &Separate &53.6 &44.2\\
       \rowcolor{gray!20} \multirow{-2}*{Cold-Start} & Join &48.2 &42.4 \\
       &Separate &61.5 &55.0 \\
      \multirow{-2}*{UniVR} & Join &63.8 &55.4 \\
\end{tabular}
\end{minipage}
}
\hfill
\hspace{0.04\linewidth}%
\subfloat[
Evaluation on other benchmark data.
\label{tab:abla_otherbench}
]{
\begin{minipage}[t]{0.64\linewidth}
\centering
\tablestyle{4pt}{1.08}
\footnotesize
\begin{tabular}{l c c c}
    Method & WorldArena~\cite{shang2026worldarena} & Uni-MMMU~\cite{zou2025uni} &RBench~\cite{deng2026rethinking} \\
    \shline
    \specialrule{0em}{0pt}{1pt}
      Emu3.5 &40.3 &28.7  &31.2\\
      Cold-Start &42.4 &37.6 &36.8 \\
      UniVR  &49.5 &54.4  &47.7 \\
      $\triangle$\textit{v.s.} Emu3.5 & \textcolor{ForestGreen}{$\uparrow$ 9.2} & \textcolor{ForestGreen}{$\uparrow$ 25.7} & \textcolor{ForestGreen}{$\uparrow$ 16.5} \\

\end{tabular}
\end{minipage}
}
\hfill
\caption{\textbf{Ablation studies.} ``LP'' and ``GR'' denote long-term planning and general reasoning in VR-X, respectively. }
\label{tab:ablations} 
\end{table*}

\noindent \textbf{Unified generation models.}
Rows 9-16 evaluate various vision-language unified models. Except for Emu3.5, most lack contiguous image generation and must iteratively unroll for long-horizon planning, generally underperforming with a mere 30\% peak success rate. This likely stems from training objectives centered on entertainment generation and artistic editing rather than logical reasoning. Emu3.5 achieves the highest score in this group, benefiting from native pre-training on diverse daily-life and handcrafting sequences. Yet textual reasoning remains indispensable for its execution, and its performance stagnates at 35\% when confronted with VR-X's intricate visual dynamics. Meanwhile, these models also exhibit poor JEPA scores, lagging significantly behind sequences rendered by text-based MLLMs, with feature distribution divergences from ground truth approximately 2–4.5$\times$ larger. This underscores that visual space reasoning remains a critical bottleneck for existing models.

\noindent \textbf{UniVR.}
In contrast, UniVR significantly enhances performance across all tasks. Despite the absence of fine-grained text procedural annotations, directly training in visual space achieves a 60\% success rate in long-term planning and 70\% in general reasoning. Remarkably, at a 34B parameter scale, UniVR outperforms Gemini 2.5 Pro on several key metrics. This shows that our training pipeline effectively empowers the model to distill essential task-centric policies directly from visual signals, bypassing the need for textual mediation. Furthermore, by fostering stronger visual reasoning, the generated sequences exhibit enhanced physical
dynamics, leading to better JEPA metrics.

\subsection{Ablation Study}
\label{subsec:abl}

\noindent \textbf{Visual reasoning benefit multimodal  understanding.} Tab.~\ref{tab:abla_understand} presents an ablation on standard multimodal understanding benchmarks. Following the protocol in~\cite{luo2026torchumm}, row one reports the baseline performance of Emu3.5 on benchmarks such as MMMU~\cite{yue2024mmmu} and MME~\cite{fu2023mme}, where the model is prompted to autonomously generate intermediate visual reasoning images and the final textual answer. In row two, we train solely on textual reasoning chains derived from VR-X (details of texual reasoning annotation provided in Appendix~\ref{sec:appA}) to align with the conventional training paradigm of multimodal understanding models, yet this yields negligible gains.In contrast, integrating the UniVR training paradigm (row three) leads to more significant gains across all six metrics. This indicates that enhanced visual reasoning serves as a powerful complement to textual supervision, effectively bolstering overall multimodal comprehension.

\noindent \textbf{Reward in VR-GRPO.} Tab.~\ref{tab:abla_vrgrpo} ablates the individual components of VR-GRPO. Row one establishes the baseline performance of the cold-start model before RL. While the global reward alone improves general reasoning (short sequences) by 3.6\% , it causes degradation in long-term planning and JEPA scores, indicating the occurrence of reward hacking. Conversely, introducing the step-focal reward alone shows positive improvements across all three metrics. Combining both rewards results in a synergistic effect, further enhancing overall performance. Finally, transitioning the VLM from absolute scalar scoring to a pairwise comparison protocol further enhances the robustness of the rewards and improves the final results.

\noindent \textbf{VR-GRPO is compatible with text-based RL.}
Tab.~\ref{tab:abla_modality} investigates the compatibility of VR-GRPO with existing multimodal rewards. In the cold-start phase, adding textual reasoning chains (row 2) yields no significant gains over pure visual reasoning (row 1). Furthermore, incorporating aesthetic-centric rewards like HPSv3 or vision-language alignment via CLIP (rows 3-4) provides limited assistance for tasks requiring complex visual reasoning, with HPSv3 showing negligible impact. In contrast, VR-GRPO (row 5) significantly boosts visual reasoning capabilities and remains compatible with other reward functions, highlighting the generalizability and versatility of our approach.

\noindent \textbf{VR-GRPO stabilizes training with heterogeneous knowledge.} Long-term planning and general reasoning in VR-X exhibit significant disparities in terms of temporal scale, environmental context, visual appearance, and knowledge domains. Rows 1-3 in Tab.~\ref{tab:abla_heter} investigate the joint learning capabilities across these tasks without the incorporation of VR-GRPO. The results indicate that during the pure cold-start phase, joint training struggles to effectively balance heterogeneous task knowledge, failing to surpass the performance of models trained individually on each task. In contrast, the introduction of VR-GRPO in row 4 empowers the model to autonomously explore optimal policies, yielding performance gains in both categories. This demonstrates the efficacy of our approach in facilitating unified visual reasoning training.

\noindent \textbf{Evaluate on other visual reasoning benchmarks.}
To further assess UniVR's generalization, we evaluate it on test data from three external benchmarks (Tab.~\ref{tab:abla_otherbench}): two focusing on embodied intelligence scenarios~\cite{shang2026worldarena,deng2026rethinking} and one on cognitive reasoning~\cite{zou2025uni}. Using the instructions and query images from these benchmarks, we prompt UniVR to generate corresponding visual reasoning trajectories, then evaluate them with the same VLM scoring protocol as in Sec.~\ref{sec:exp}. Despite significant domain gaps from our training data, encompassing unseen robotic manipulation tasks, environments, camera viewpoints, and mathematical/physical reasoning questions—UniVR substantially improves baseline performance (up to 24.3\%) on these new tests. Moreover, with VR-GRPO, this gain significantly exceeds that of vanilla SFT, demonstrating that VR-GRPO fosters more generalizable visual reasoning.

\section{Conclusion}
In this work, we presented UniVR, a unified framework that enables complex reasoning and planning directly within visual space, free from dense language supervision. Central to our approach is VR-GRPO, which enforces both global task completion and fine-grained step-level physical coherence through complementary rewards. Trained and evaluated on the diverse VR-X benchmark, UniVR achieves substantial gains over strong text-based reasoning pipelines and unified generation models, demonstrating that raw visual demonstrations alone can support sophisticated, long-horizon policy learning. Beyond task execution, our findings reveal that strengthened visual reasoning confers broader benefits to multimodal understanding, suggesting vision-native computation as a powerful complement to textual abstraction. We believe this work opens promising avenues toward more grounded, efficient world models that learn directly from the abundant visual world.

\clearpage

\bibliographystyle{plainnat}
\bibliography{main}

\clearpage

\beginappendix

In this supplementary appendix, we provide additional details and analyses:
\begin{itemize}
    \item Appendix~\ref{sec:appA} covers implementation details of the two-stage training pipeline
    \item Appendix~\ref{sec:appB} presents further analysis of the VR-X benchmark and evaluation metrics
    \item Appendix~\ref{sec:appC} includes additional ablation studies and visualizations.
\end{itemize}

\section{Implementation Details}
\label{sec:appA} 

\subsection{Cold Initialization}
\begin{figure}[t]
    \centering
    \includegraphics[width=0.9\linewidth]{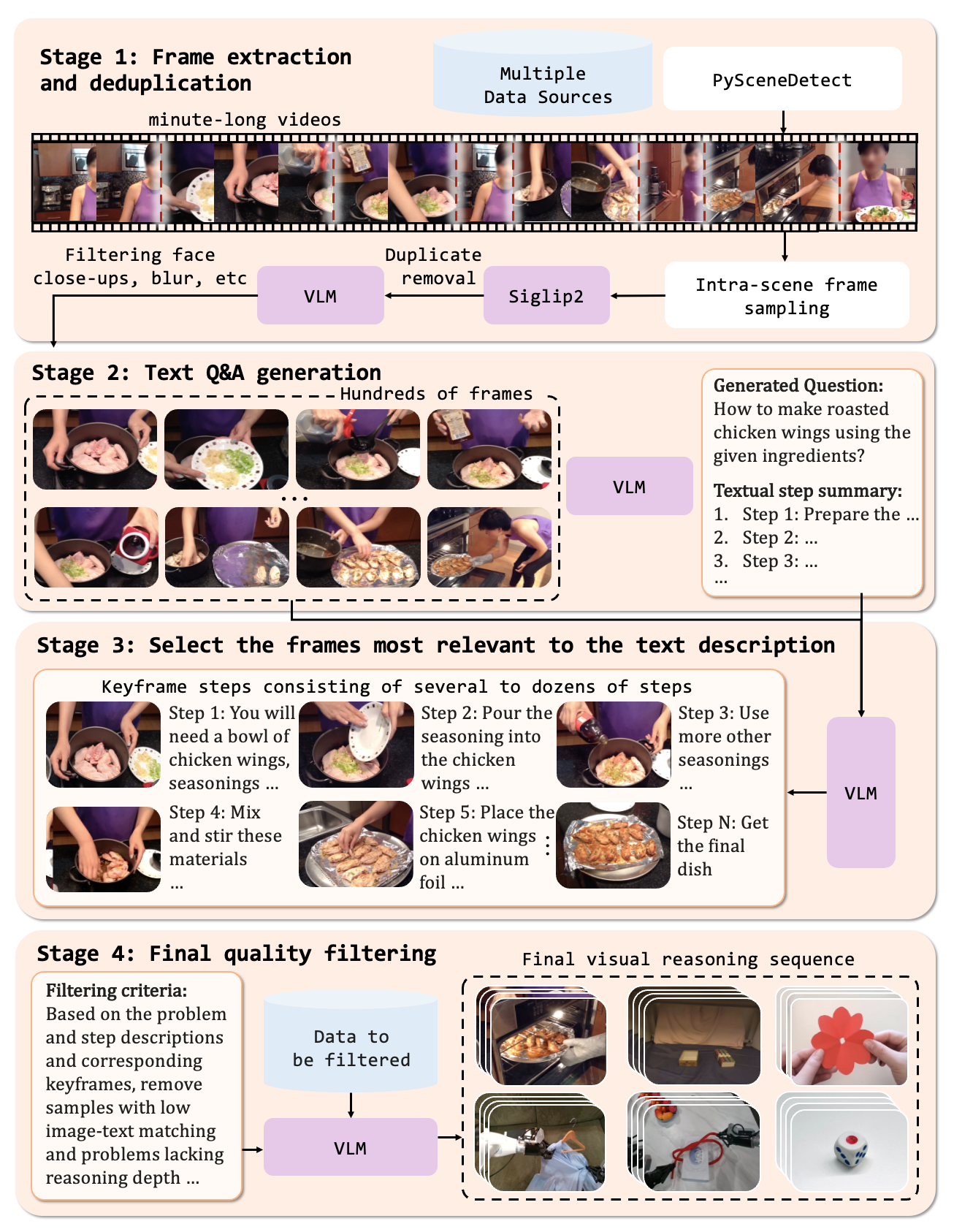}
    \caption{\textbf{Data Generation Pipeline.}  }
    \label{fig:data_gen}
\vspace{-1.5em}
\end{figure}

\noindent \textbf{Data Construction.} We construct our SFT training data from multiple sources, as summarized in Tab.~\ref{tab:supp_training_data}. These datasets vary in size and diversity; we weight all subsets according to their sample counts during training. Using the pipeline shown in Fig.~\ref{fig:data_gen}, we filter raw data into curated training samples. This data processing pipeline consists of four stages. First, we select raw video sequences from the source data and perform scene-aware temporal sampling via PySceneDetect~\cite{castellano2025pyscenedetect} at 0.27 FPS, which preserves richer information than random sampling. This is followed by SigLIP2-based~\cite{tschannen2025siglip} deduplication and VLM-based filtering to remove low-quality frames such as facial close-ups, blurred frames, and blank screens, typically yielding hundreds of frames from a minute-long video. Second, we leverage Qwen3.5-397B~\cite{qwen3.5} to synthesize reasoning-oriented questions and corresponding textual answers from the sampled sequences, with approximately 10 key steps per trajectory. Third, conditioned on these QA pairs, we prompt Qwen3.5 to identify the most relevant query image and key-step frames. Finally, we apply a rigorous quality filter to discard low-quality samples, including those with image-text mismatches, trivial questions requiring no reasoning or planning, and poor visual quality. This pipeline filters out nearly 80\% of the candidate pool, ensuring that the remaining samples represent the most informative and logically sound visual reasoning trajectories.

For non-video sources such as VisualCoT and ZebraCoT, which are already formatted as image sequences, we bypass temporal sampling and standardize them to match video-derived data. Overall, we curate 310k cold initialization samples from 1.5M raw candidates.

\noindent \textbf{Optimization.} For cold initialization, we initialize from Emu3.5-34B and train UniVR using the configurations in Tab.~\ref{tab:supp_detail} (second column) on 32 GPUs with full parameters. Each image is resized to 512 on the short side and tokenized by Emu3.5's VQ tokenizer, yielding 1,000–1,500 tokens per image. We cap the maximum sequence length per sample at 15,000 to reduce training overhead while accommodating reasoning trajectories spanning approximately several minutes.

\begin{figure}
    \centering
    \includegraphics[width=\linewidth]{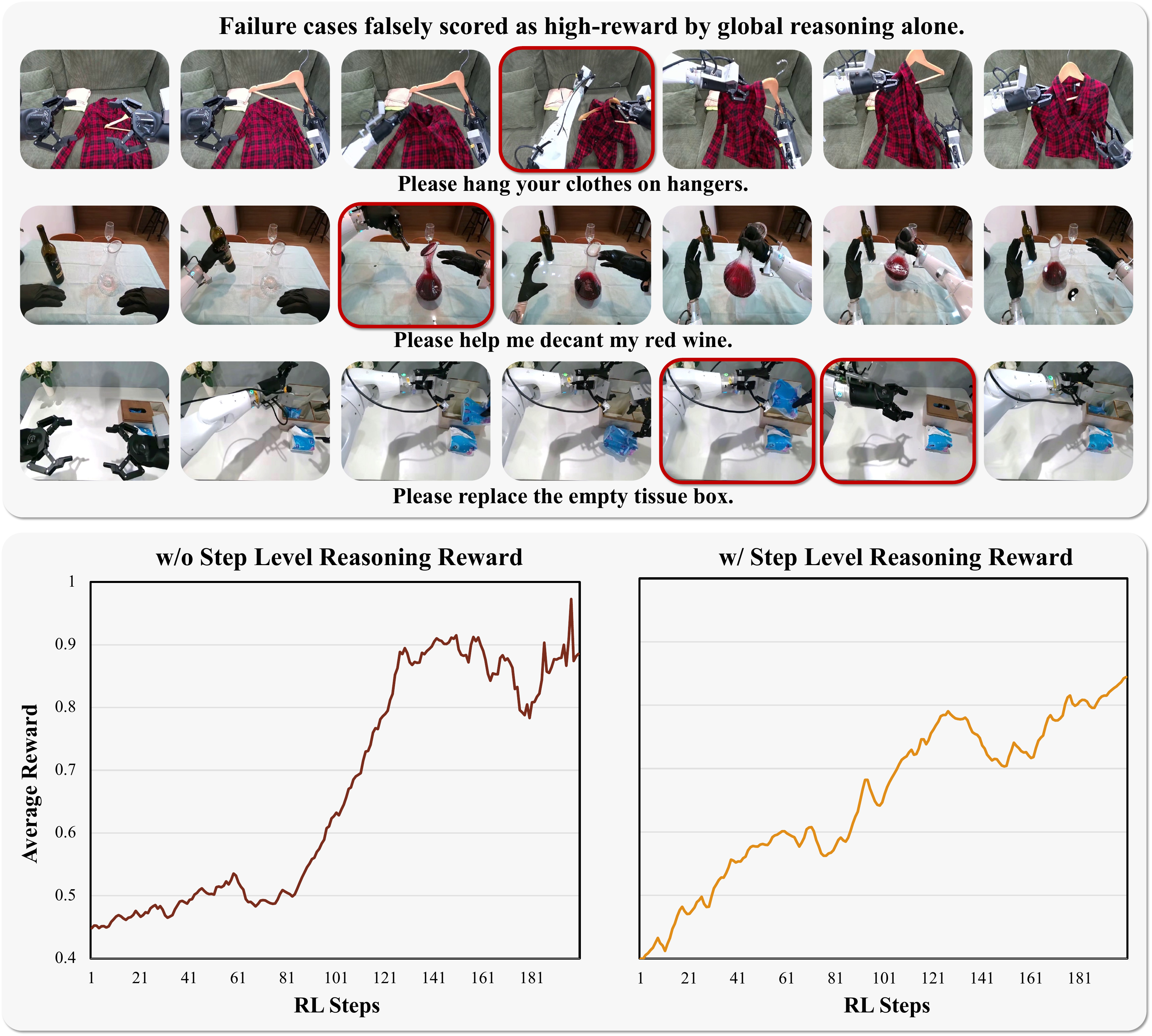}
    \caption{\textbf{Analysis of the Visual Reasoning Reward.} (Up) Relying solely on the global reward misses step-level errors in long-horizon planning tasks, \textit{e.g.} hangers penetrating garments (top row), erroneous liquid pouring dynamics (middle row), and jittering transitions between paper towels (bottom row). (Down) RL reward curves with and without the step-level reward component. }
    \label{fig:reward_comp}
\vspace{-1em}
\end{figure}

\subsection{Reinforcement Learning}
\noindent \textbf{Data Construction.} UniVR employs a novel visual reasoning reward that requires sequences with complex reasoning procedures. To this end, we filter hard samples from the original 310k training set using the post-initialization model, yielding approximately 3k samples. This subset comprises roughly 2k long-term planning trajectories, predominantly spanning 6–10 reasoning steps—and 1k general reasoning data to preserve task diversity.

\begin{table}[t]
\centering
\caption{Training hyperparameters for Cold-Start and RL stages.}
\label{tab:supp_detail}
\begin{tabular}{l|c|c}
Hyperparameters & Cold-Start & RL \\
\shline
    \specialrule{0em}{0pt}{1pt}
Learning rate &5$\times$10$^{-4}$ &1$\times$10$^{-5}$ \\
LR scheduler &Cosine &Cosine \\
Weight decay &0.1 &0.1 \\
Gradient norm clip &5.0 &5.0 \\
Warm-up steps &700 &0 \\
Training steps &10k & 250 \\
Sequence length &15000 &15000 \\
Batch Size &128 &128 \\
Resolution &[512, 640] &[512, 640] \\
$\lambda$ &- &2.0 \\
\end{tabular}
\vspace{-1em}
\end{table}

\begin{table}[h]
\centering
\caption{Data composition.}
\label{tab:supp_training_data}
\begin{tabularx}{\textwidth}{l|c|>{\centering\arraybackslash}X|>{\centering\arraybackslash}X}
Category & Data Source & Frames & Ratios \\
\shline
\specialrule{0em}{0pt}{1pt}
\multirow{4}{*}{Visual Guidance} & EgoDex~\cite{hoque2025egodex} & 289,053 & 23.7\% \\
 & Action100M~\cite{chen2026action100m} & 109,029 & 5.0\% \\
 & Epic-kitchen~\cite{damen2020epic} & 53,605 & 2.5\% \\
 & VideoCraftBench~\cite{ren2026videoworld} & 2,272 & 1.0\% \\
\midrule
\multirow{4}{*}{Robot Manipulation} & AgiBot~\cite{bu2025agibot} & 427,267 & 24.5\% \\
 & Droid~\cite{khazatsky2024droid} & 13,000 & 1.0\% \\
 & Bridge~\cite{walke2023bridgedata} & 14,850 & 1.6\% \\
 & ZebraCoT-Robot~\cite{li2025zebra} & 54,270 & 3.5\% \\
\midrule
Editing & ZebraCoT-Multiobject~\cite{li2025zebra} & 128,075 & 6.5\% \\
\midrule
\multirow{2}{*}{Spatial Perception} & ThinkMorph-Navigation~\cite{gu2025thinkmorph} & 68,568 & 11.2\% \\
 & ZebraCoT-Embodiment~\cite{li2025zebra} & 58,931 & 3.2\% \\
\midrule
\multirow{2}{*}{Visual Search} & VisualCoT~\cite{shao2024visual} & 30,000 & 4.9\% \\
 & ThinkMorph-Search~\cite{gu2025thinkmorph} & 13,980 & 2.3\% \\
\midrule
\multirow{3}{*}{Puzzle \& Game} & VRBench~\cite{yang2025vrbench} & 13,000 & 1.0\% \\
 & Zebra-Jigsaw~\cite{li2025zebra} & 43,798 & 7.1\% \\
 & ThinkMorph-VisPuzzle~\cite{gu2025thinkmorph} & 13,000 & 1.0\% \\
\end{tabularx}
\vspace{-1em}
\end{table}

\noindent \textbf{Optimization.} For reinforcement learning, we initialize from the SFT checkpoint and train with full parameters on 32 GPUs using the configuration in Tab.~\ref{tab:supp_detail} (third column). The Qwen evaluator is deployed on an additional 8 GPUs. To improve training efficiency, we use Qwen3-VL-30B-A3B as the evaluator. Larger variants (e.g., Qwen3-VL-235B or Qwen3.5-397B) yield more accurate rewards but significantly increase training latency. Classifier-Free Guidance (CFG) is disabled during rollout.

\section{Details on VR-X Benchmark}
\label{sec:appB}
The VR-X evaluation set comprises 1.8k high-quality reasoning trajectories curated by professional annotators from a held-out subset (detailed distribution in Tab.~\ref{tab:supp_training_data}). VR-X employs two metrics: VLM score and JEPA similarity score, detailed below.

\noindent \textbf{VLM Evaluation Detail.}  For the VLM score, we design a unified prompt covering visual quality, task completion, logical coherence, physical dynamics, and temporal consistency. Both the GT and generated sequence are fed into the VLM, enabling it to reference the GT logic process and action dynamics for more accurate judgment. To assess the alignment between VLM and human evaluation, we first have professional annotators score model outputs on the benchmark. The samples are then shuffled and presented with corresponding VLM and human scores, allowing annotators to judge which is superior. From this, we compute the Spearman correlation~\cite{spearman1904proof} between human and VLM scores, which reaches approximately 0.85, indicating a high degree of human alignment.

\noindent \textbf{JEPA Evaluation Detail.} For the JEPA score, we follow the implementation in~\cite{luo2024beyond}. Specifically, its computation resembles traditional video metrics such as FVD~\cite{unterthiner2018towards}, but replaces the I3D~\cite{carreira2017quo} feature extractor with a V-JEPA~\cite{assran2025vjepa2} encoder. Image sequences are compressed into 1280-dimensional latent vectors, and the distance between two feature distributions is computed via Maximum Mean Discrepancy (MMD) with a polynomial kernel—smaller distances indicate higher similarity. Since the V-JEPA encoder is trained to encode spatiotemporal coherence and physical dynamics, the JEPA score serves as a complement to the VLM score, assessing whether generated sequences conform to real-world transitions. This metric is applied only to the long-term planning subset, as it targets visual sequences containing real-world dynamics and is unsuitable for single-step reasoning tasks or simulated scenarios in general reasoning. Compared to FVD, the JEPA score converges with approximately 1,000 samples, aligning with the scale of the VR-X benchmark.

\section{More Analysis and Visualizations}
\label{sec:appC}
\noindent \textbf{Visual reasoning rewards mitigates reward hacking.} Fig.~\ref{fig:reward_comp} presents additional cases where the global reasoning reward alone yields near-perfect VLM scores, yet manual inspection reveals physical and logical errors, such as implausible hanger-garment interactions, incorrect wine-pouring dynamics, and flawed towel-replacement logic. These samples span extended durations (30+ seconds), with errors localized to merely a few frames that global VLM assessment easily overlooks. This phenomenon is reflected in the training curves (Fig.~\ref{fig:reward_comp}, bottom left) the global-reward curve exhibits sharp spikes and top-end oscillations, indicating shortcut-seeking behavior that neglects intermediate-step correctness. In contrast, the VR-GRPO curve (bottom right) ascends more smoothly, demonstrating substantially lower reward hacking risk.
\leavevmode
\begin{wraptable}[8]{r}{0.5\linewidth}
    \centering
    \tablestyle{4pt}{1.1}
    \begin{tabular}{l | cccc}
        Method & <10s & 10-30s & 30-60s & >60s \\
        \shline
        \specialrule{0em}{0pt}{1pt}
        Emu3.5      &54.2 &48.0 &38.9 &21.7 \\
        Cold-Start  &58.7  &54.5  &41.1  &26.9 \\
        UniVR       &71.1  &61.0  &56.7  &45.6 \\
        $\triangle \textit{v.s.}$ Emu3.5   & \textcolor{ForestGreen}{$\uparrow$ 16.9} & \textcolor{ForestGreen}{$\uparrow$ 13.0} & \textcolor{ForestGreen}{$\uparrow$ 17.8} & \textcolor{ForestGreen}{$\uparrow$ 23.9} \\
    \end{tabular}
    \caption{Performance across different time span.}
    \label{tab:duration_ablation}
\end{wraptable}
\noindent \textbf{VR-GRPO stabilizes long-horizon visual reasoning.} We analyze the scalability of VR-GRPO across varying temporal horizons by partitioning test samples into four duration-based groups: <10s, 10–30s, 30–60s, and >60s. As shown in Tab.~\ref{tab:duration_ablation}, VR-GRPO delivers the most pronounced gains on >30s sequences. We attribute this to its step-level reward design, which explicitly maintains logical coherence and physical consistency at intermediate stages. By correcting error-prone steps, VR-GRPO effectively mitigates compounding errors that typically accumulate during long-range prediction, thereby exhibiting superior stability on extended reasoning traces.

\begin{figure}
    \centering
    \includegraphics[width=0.9\linewidth]{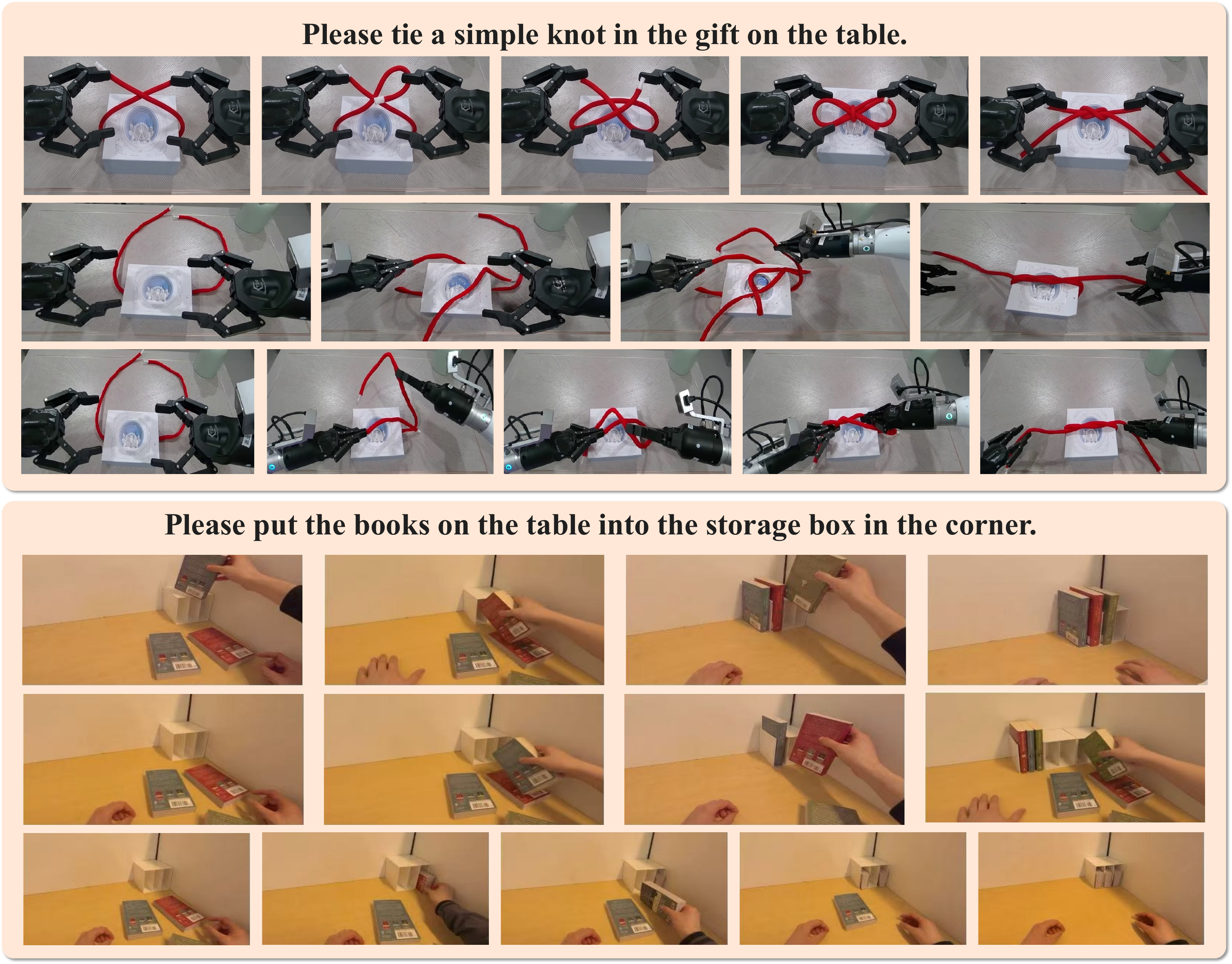}
    \caption{\textbf{Comparison with Gemini and Emu3.5.} In each group, the first, second, and third rows correspond to Gemini 3 Pro + Nano Banana 2, Emu3.5, and UniVR, respectively.}
    \label{fig:failure_case}
\end{figure}

\noindent \textbf{More Comparisons with Baselines.} Fig.~\ref{fig:failure_case} provides additional side-by-side comparisons with Gemini 3 Pro + Nano Banana 2 and Emu3.5 on identical test samples. While both baselines generate high-fidelity visual appearances, they exhibit execution inaccuracies in fine-grained tasks such as rope manipulation, garment unfolding and folding, paper-folding dynamics, and environmental consistency during policy execution. Benefiting from our visual reasoning training pipeline, UniVR achieves comparable visual fidelity while demonstrating superior logical coherence and physical consistency.

\begin{figure}
    \centering
    \includegraphics[width=0.9\linewidth]{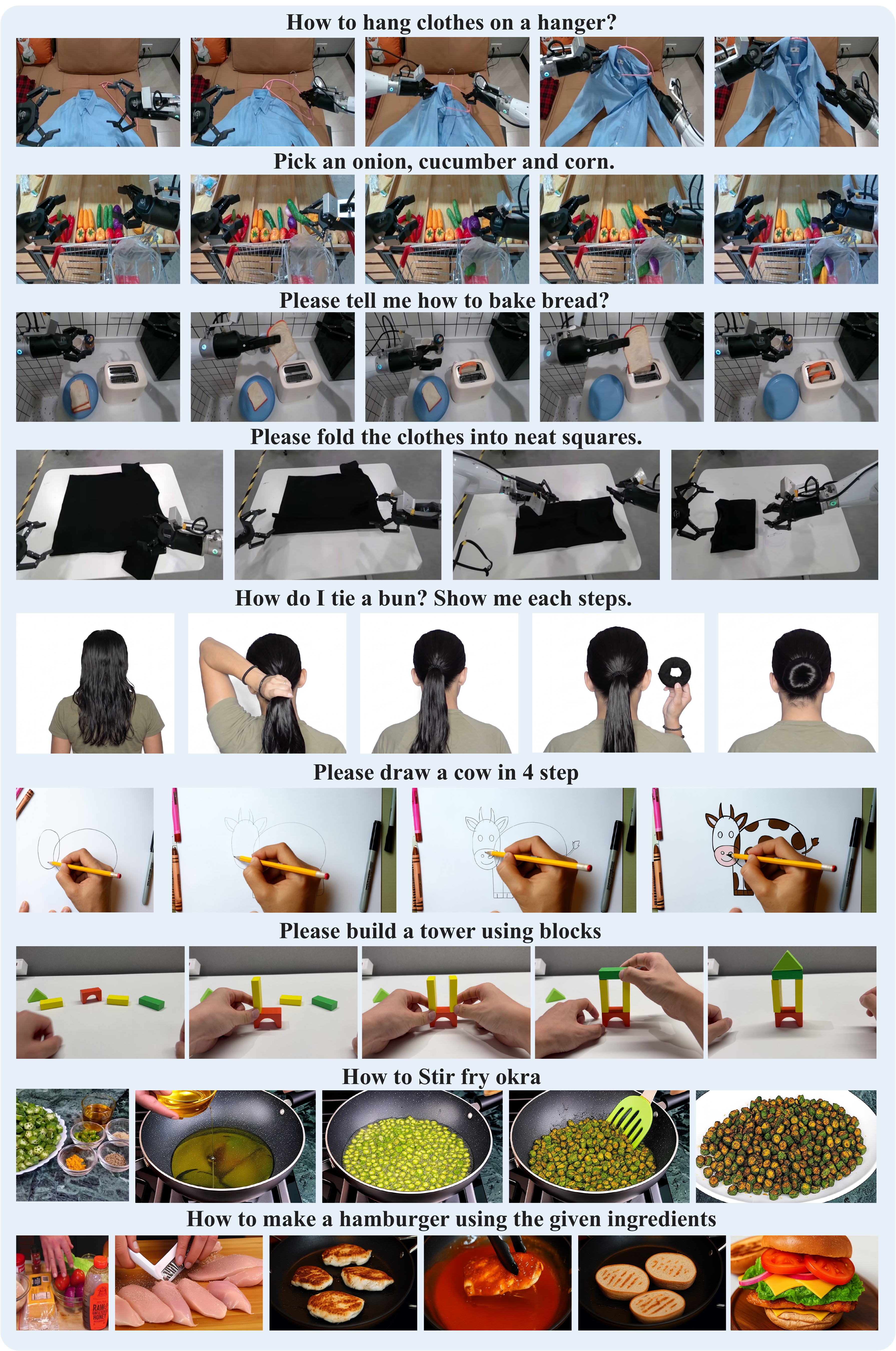}
    \caption{\textbf{More visualization of UniVR.} }
    \label{fig:more_vis}
\vspace{-1.5em}
\end{figure}

\noindent \textbf{More Visualization.}  Fig.~\ref{fig:more_vis} shows results of UniVR across additional scenarios, generating long-horizon sequences with coherent logic, physical dynamics, and temporal consistency.

\noindent \textbf{Limitations.} Despite these promising results, UniVR presents several limitations worth noting. First, our training on 34B-scale models with long visual sequences demands substantial computational resources, potentially limiting accessibility. Second, although VR-GRPO improves evaluation quality via its step-level design, our reward mechanism still relies on general purpose VLMs with limited fine-grained physical-world knowledge. A more powerful reward system natively grounded in visual world dynamics is needed. Finally, although visual reasoning holds significant potential for learning complex world knowledge and enhancing multimodal understanding, achieving optimal synergy among native visual, textual, and auditory reasoning remains an open question for future work.

\end{document}